\documentclass[twocolumn,10pt]{article}

\usepackage{graphicx}
\usepackage{amsmath}
\usepackage{booktabs}
\usepackage{hyperref}
\usepackage{caption}
\usepackage{subcaption}
\usepackage{url}
\usepackage{xcolor}
\usepackage{float}
\usepackage{placeins}  

\usepackage[margin=1in]{geometry}

\usepackage{rotating}

\usepackage[numbers]{natbib}

\title{Color histogram equalization and fine-tuning to improve expression recognition of (partially occluded) faces on sign language datasets}

\author{Fabrizio Nunnari, Alakshendra Jyotsnaditya Ramkrishna Singh, Patrick Gebhard \\
fabrizio.nunnari@dfki.de, alsi04@dfki.de, patrick.gebhard@dfki.de \\
 German Research Center for Artificial Intelligence (DFKI)}	

\date{}

\begin{document}

\maketitle

\begin{abstract}
The goal of this investigation is to quantify to what extent computer vision methods can correctly classify facial expressions on a sign language dataset. We extend our experiments by recognizing expressions using only the upper or lower part of the face, which is needed to further investigate the difference in emotion manifestation between hearing and deaf subjects.
To take into account the peculiar color profile of a dataset, our method introduces a color normalization stage based on histogram equalization and fine-tuning.
The results show the ability to correctly recognize facial expressions with 83.8\% mean sensitivity and very little variance (.042) among classes. Like for humans, recognition of expressions from the lower half of the face (79.6\%) is higher than that from the upper half (77.9\%). Noticeably, the classification accuracy from the upper half of the face is higher than human level.
\end{abstract}

\subsection*{Keywords}
sign language, emotion, partial face visibility, Ekman, facial expression, histogram equalization, color normalization.


%
%
%
\section{Motivation and related work}

\begin{figure*}
  \centering
  \includegraphics[width=0.115\textwidth]{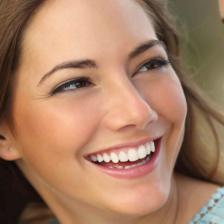}\hspace{1pt}%
  \includegraphics[width=0.115\textwidth]{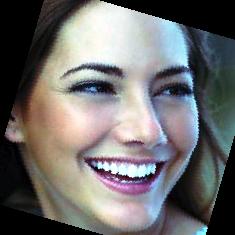}\hspace{1pt}%
  \includegraphics[width=0.115\textwidth]{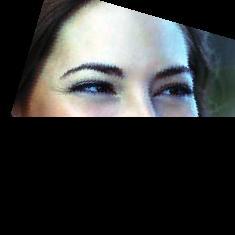}\hspace{1pt}%
  \includegraphics[width=0.115\textwidth]{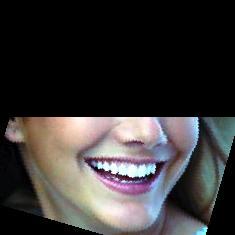}\hspace{1pt}%
  \includegraphics[width=0.115\textwidth]{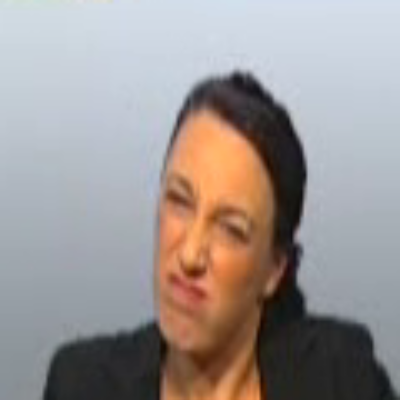}\hspace{1pt}%
  \includegraphics[width=0.115\textwidth]{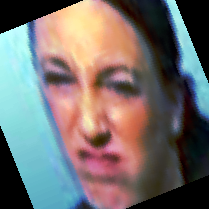}\hspace{1pt}%
  \includegraphics[width=0.115\textwidth]{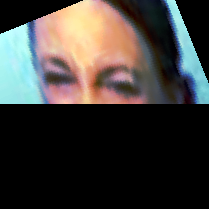}\hspace{1pt}%
  \includegraphics[width=0.115\textwidth]{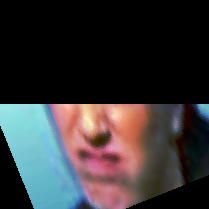}%
  \caption{Example pictures of our procedure to classify Ekman emotions. The original picture is first normalized in size, rotation, and color profile. Then, we perform classification on the full face, only top part, and only bottom part. On the left: examples from the AffectNet dataset. On the right: examples from the PHOENIX sign language dataset.}
  \label{fig:teaser}
\end{figure*}


Research in emotion expression in sign language (SL) is receiving a lot of attention from the research community. For example, see the recent workshop on Expressing Emotions in Sign Language\footnote{ExEmSiLa - \url{https://www.idgs.uni-hamburg.de/en/forschung/tagungen/expressing-emotions-in-sign-languages.html}}. From an overview of the reported work, it seems that, to conduct studies, the linguistic community mainly relies on manual observation and annotation.

For example, Kirst (2024) \cite{kirst_eyebrow_2024} reports on an experiment aiming to study the “layering” as “when multiple phonological and paralinguistic elements co-occur”, and focuses on the use of eyebrows for the communication of emotions. The study is conducted through an interpretation of manually annotated FACS (Facial Action Coding System) units, which is a very time-consuming task which could be automated using computer vision libraries.

Kulshreshtha (2024) \cite{kulshreshtha_effect_2024} found that, in ISL, participants used the doubled-wh signs in the context of anger, surprise, and excitement. The analysis was performed manually, to discover a pattern that could be identified by unsupervised machine learning approaches.\\

In this paper, we present the beginning of an investigation aiming to understand to what extent machine learning, and in particular computer vision based on convolutional neural networks, can help in further investigating emotion recognition strategies in sign language.
More specifically, we report the results of an experiment that aims to correctly classify emotions based on the classification of Ekman expressions~\cite{ekman_constants_1971}.

To further connect this investigation to the realm of sign language, we followed the observations that, when judging emotions, native signers and speaking people pay attention differently to the top and bottom parts of the faces.

For example, Letourneau and Mitchell \cite{letourneau_gaze_2011} measured, through eye gaze analysis, how much hearing and deaf individuals fixate faces when recognizing identities and emotions. It seems that, w.r.t. hearing people, deaf signers focus more on the bottom half of the faces when judging emotions.
The same results were recently replicated by Kang et al. \cite{kang_emotion_2025}.

Dye et al.~\cite{dye_response_2016} also confirmed for other ``visuospatial selective attention tasks [...] a shift toward attending to the inferior visual field in users of a signed language'' for ASL signers.\\

We tested our facial expression recognition (FER) machine learning pipeline on two datasets. The first is AffectNet \cite{mollahosseini_affectnet_2019}: a popular big-sized facial expression dataset widely used to perform FER tasks. The goal being to quantify to what extent emotions can be correctly classified using only a half of the image w.r.t. the full frame.
The second dataset is FePh (Facial Expression PHOENIX) \cite{alaghband_facial_2021}. This is a subset of the popular PHOENIX dataset \cite{camgoz_neural_2018} where 3300 samples were annotated for facial expression.
The tests on this second subset aim at observing if, when applied to a sign language facial expression dataset, FER follows the same pattern w.r.t. a generic facial expression dataset.\\

This paper presents three contributions.
First, we introduce an image pre-processing technique for FER tasks based on cropping, squaring, zooming, normalizing color, and rotating. In particular, we propose a color normalization stage (using Histogram Equalization) that decouples implementation from hard-coded RGB median shifts and improves performances among expressions. Histogram equalization as an image preprocessing tool has already been successfully applied in other contexts (e.g., \cite{pontalba_assessing_2019,pei_robustness_2023}). To our knowledge, this is the first application in the FER and sign language domains.

Second, we quantify the ability of computer vision algorithms in recognizing emotions using only a half of the face. This second result suggests that there is the possibility of performing emotion recognition even in scenarios where half of the face is covered (e.g., while wearing hygienic masks or virtual reality helmets).

Third, we apply our method to a sign language dataset, showing that our image pre-processing and training pipeline increase significantly FER performances with respect to the state of the art, and that the same emotion recognition pattern emerges when expressing emotions while signing.

\section{Method}

\subsection{Datasets preparation}

To run our experiments, we relied on two datasets, whose labels distribution is reported in table \ref{tab:datasets-samples}.

\textbf{AffectNet} \cite{mollahosseini_affectnet_2019} is one of the most popular datasets for FER tasks, which consists of 800k images, gathered from the internet and labeled by online participants on the seven Ekman expressions plus \textit{Neutral}, \textit{Other}, and \textit{Uncertain} labels.
Of all images, about 420,000 images were manually annotated, and the rest automatically annotated using a ResNet neural network.

For our experiments we include only the manually annotated images of the 7 Ekman emotions plus Neutral, removing images with ``unknown'' or multiple labels, resulting in a total of 273562 training images and 3999 validation set images, containing about 500 images from each class.

To test our method on the sign language domain, we used the \textbf{Facial Expression PHOENIX (FePh)} dataset \cite{alaghband_facial_2021}; possibly the only dataset which associates sign language faces with affect detection labels.
It consists of 3359 frames, extracted from the PHOENIX dataset \cite{forster_rwth-phoenix-weather_2012}, which is in turn, so far, one of the most popular datasets used for the development of systems for the recognition of German sign language \cite{camgoz_neural_2018}.

The FePH is annotated for the seven Ekman expressions, plus extra labels for \textit{neutral} and \textit{other}.
For this work, we removed the frames annotated as \textit{other} or to multiple classes.
As a result, we used only 2547 frames with a unique label.

\begin{table*}[t]
    \centering
    \footnotesize
\begin{tabular}{l|rrrrrrrr|r}
\toprule
\textbf{Dataset} & \multicolumn{1}{l}{\textbf{Happiness}} & \multicolumn{1}{l}{\textbf{Sadness}} & \multicolumn{1}{l}{\textbf{Surprise}} & \multicolumn{1}{l}{\textbf{Fear}} & \multicolumn{1}{l}{\textbf{Anger}} & \multicolumn{1}{l}{\textbf{Disgust}} & \multicolumn{1}{l}{\textbf{Contempt}} & \multicolumn{1}{l|}{\textbf{Neutral}} & \multicolumn{1}{l}{\textbf{Total}} \\
\midrule
AffectNet Train & 134415 & 25459 & 14090 & 6378  & 24882 & 3803  & 3751  & 74874 & 273562 \\
AffectNet Val & 500   & 500   & 500   & 500   & 500   & 500   & 499   & 500   & 3999 \\
\midrule
FePH  & 179   & 349   & 818   & 308   & 507   & 187   & -     & 199   & 2547 \\
\bottomrule
\end{tabular}%
    \caption{The number of samples for each class in our training/testing datasets.}
    \label{tab:datasets-samples}
\end{table*}

\subsection{Image pre-processing}
\label{sec:img-pre-proc}

Before being used for training or prediction, images of both the AffectNet and FePh datasets are pre-processed via the following steps:

\begin{enumerate}
    \item Crop face bounding box and identify eye centers and nose tip positions using the de Paz Centeno implementation\footnote{\url{https://github.com/ipazc/mtcnn}} of the MTCNN method \cite{zhang_joint_2016};
    \item extend the shortest boundary in order to have a squared image;
    \item zoom out the face view to have a boundary edge at 110\% w.r.t. the previous step;
    \item perform color normalization via Histogram Equalization;
    \item rotate the image around the nose to bring the eyes on the same horizontal level.
\end{enumerate}

Steps 1 and 5 are reproduced from the method already proposed by Savchenko \cite{savchenko_facial_2021}.

Step 2 was introduced from the observation that, in many images, thin oblong faces lead to the detection of rectangular areas with a high ratio between edge lengths. As such, some face images would be heavily distorted when scaled to a square before being processed by the convolutional neural network. Here, we prevent distortions by preparing a squared input.

Step 3 was introduced from the observation that, in many images, the chin and forehead are only partially visible. Since some emotions are characterized by wrinkles on the forehead and opening of the mouth, by zooming out we ensure a better visibility of those areas.

\begin{figure*}
    \centering
    \includegraphics[width=0.25\textwidth]{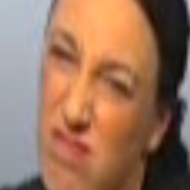}
    \hspace{0.1in}
    \includegraphics[width=0.64\textwidth]{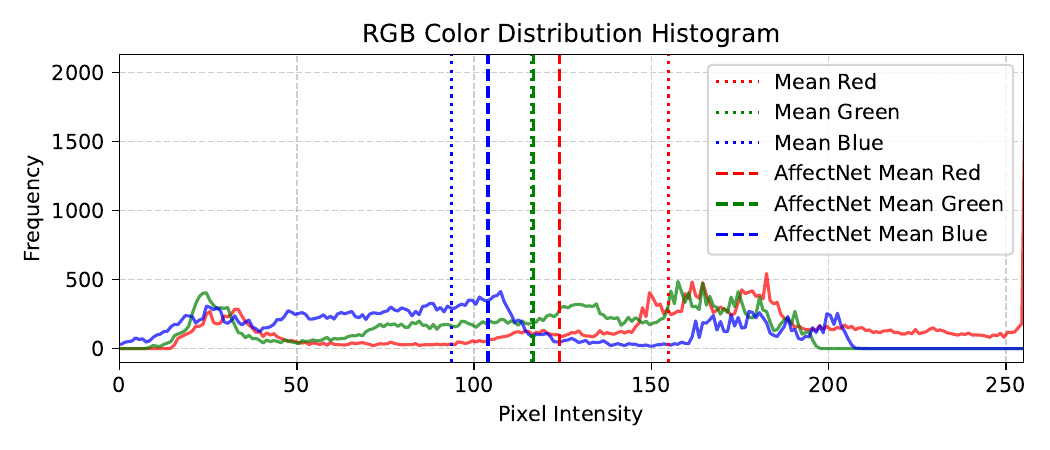}\\
    \includegraphics[width=0.25\textwidth]{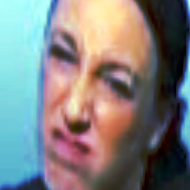}
    \hspace{0.1in}
    \includegraphics[width=0.64\textwidth]{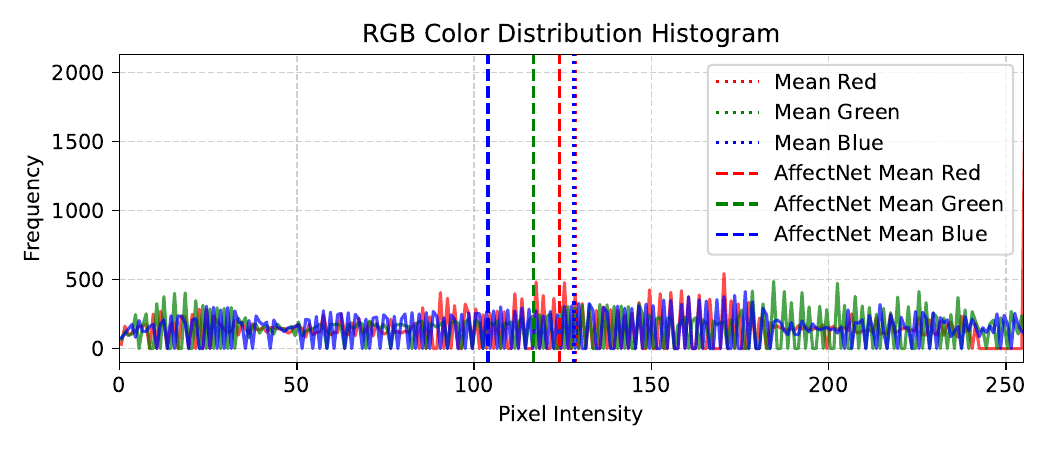}
    \caption{A sample FePh image and its color histograms. Top: the original image color histogram showing its mean values compared to the mean values of the AffectNet dataset (normally used for mean subtraction). Bottom: The color profile after Histogram Normalization.}
    \label{fig:FePh-color-histogram}
\end{figure*}

Finally, step 4 was introduced because of the intuition that facial expression recognition is in part due to the detection of shadows projected by wrinkles formed when activating facial muscles.
In the AffectNet dataset, color conditions vary considerably among images: some of them show a high contrast between bright and dark areas, while some others present a more ``flat'' color profile (that is, colors are distributed over a small portion of the possible range).
Thus, our goal is to ``stretch'' the color distribution of an image to maximize the contrast between the skin and the shadows.

In computer vision tasks that employ convolutional neural networks, the standard color preprocessing technique is the ``mean subtraction'' \cite{krizhevsky_imagenet_2012} (MS), i.e., shifting separately the R, G, and B components of an image to center them on the mean computed in the training set. This centers the data distribution, but does not modify image contrasts.

We thus introduced in our pipeline a Histogram Normalization\footnote{\url{https://en.wikipedia.org/wiki/Histogram_equalization}} \cite{russ_image_2006} procedure: a color mapping technique that reshapes the color distribution of images in order to cover the full range of pixel values and at the same time centers the mean to the middle value of the spectrum. Our guess is thus that by enhancing images contrast, performance of FER tasks will increase.

An example of Histogram Equalization is shown in Figure \ref{fig:FePh-color-histogram}. The color profile of FePh (and in general the PHOENIX dataset) is very peculiar: it does not match the RGB mean values computed on the AffectNet dataset, especially on the red channel, which is affected by a peak count at the maximum value. After applying the color normalization, the color distribution is more uniform for all three channels, means are centered at 128, and shadows are more pronounced.
As a further advantage, the Histogram Normalization procedure frees the code from hard-coded values depending on a specific training set.

It is important to perform the histogram equalization before rotating the image to avoid that the black corners formed during rotation influence the pixel distribution.

For our implementation, we used the Color Histogram Equalization implemented in OpenCV\footnote{\url{https://docs.opencv.org/4.x/d4/d1b/tutorial_histogram_equalization.html}}. The preprocessing technique described above can be reproduced by executing the script available in the DFKI Sign Language research repository\footnote{\url{https://github.com/DFKI-SignLanguage/EASIER-EkmanClassifier/blob/main/Scripts/NormalizeImages.py}}.

\subsection{Top and bottom face partitioning}

After the pre-processing, all images have been used to generate two variants: (1) top-half visible with bottom-half masked (pixels set to zero), and (2) bottom-half visible with top-half masked. The partitioning occurs at the horizontal midpoint of the image. Examples can be seen in figure \ref{fig:teaser}.

\subsection{Training and testing}
\label{sec:train-test}


As base architecture for our experiments, we employ MobileNetV2 \cite{sandler_mobilenetv2_2018} pre-trained on the ImageNet dataset \cite{krizhevsky_imagenet_2012}. The implementation uses the Pytorch\footnote{\url{https://pytorch.org}} library.

In a first step, the model is fine-tuned on the AffectNet \cite{mollahosseini_affectnet_2019} dataset for emotion classification. To perform this, we replaced existing classifier layer with our own classifier layer and trained the entire model on the AffectNet dataset. For AffectNet, we had separate train and validation sets. The training was performed on a batch size of 64, with Adam optimizer and early stopping.

In the second step, we used the model trained on AffectNet and further fine-tuned it on the FePh dataset as described by Desphande et al.~\cite{deshpande_fine-tuning_2022}. The paper suggests using a two-stage fine tuning: the first on the final classification layers, and the second on the full model. In addition, they use SAM optimizer \cite{foret_sharpness-aware_2021} and class imbalance handling.

For the FePh dataset, given the limited amount of samples, we performed all tests using a 5-fold cross-validation method.\\

We performed the above-described procedure for each of the three image conditions: full, top and bottom.
We evaluated models using multiple metrics including accuracy, balanced accuracy, and class-wise sensitivity to provide comprehensive performance assessment across all emotion categories for all face regions (full, top, and bottom).

\section{Experiments}

\begin{table*}
\begin{sideways}
    \centering
\begin{minipage}{\textheight} 
    \setlength{\tabcolsep}{3pt}  
    \footnotesize
    \begin{tabular}{rl|llll|r|cccccccc|rrr}
    \toprule
          &       &       &       &       &       &       & \multicolumn{8}{c|}{\textbf{Class sensitivities}}             & \multicolumn{1}{l}{\textbf{Bal.}} & \multicolumn{1}{l}{\textbf{Mean}} &  \\
          & \textbf{Method} & \textbf{Img proc.} & \textbf{Hyp. par.} & \textbf{Train Set} & \textbf{Test Set} & \multicolumn{1}{l|}{\textbf{Acc.}} & \textbf{Hap.} & \textbf{Sad.} & \textbf{Sur.} & \textbf{Fear} & \textbf{Ang.} & \textbf{Dis.} & \textbf{Con.} & \textbf{Neu.} & \textbf{Acc.} & \textbf{Sens.} & \textbf{SD} \\
    \midrule
    1     & Savchenko & MS, 1, 5 &       & AffectNet-Full & AffectNetVal-Full & .561 & .898 & .606 & .544 & .622 & .628 & .582 & 0     & .612 & .561 & .562 & .251 \\
    \midrule
    2     & Ours (MobileNet) & 1, 2, 3, 5 & bs64, ep20 & AffectNet-Full & AffectNetVal-Full & .549 & .696 & .632 & .468 & .486 & .684 & .516 & .535 & .372 & .549 & .549 & .113 \\
    \midrule
    3     & Ours (MobileNet) & 1-5   & bs64, ep20 & AffectNet-Full & AffectNetVal-Full & .571 & .778 & .532 & .546 & .644 & .550 & .504 & .419 & .594 & .571 & .571 & .106 \\
    4     & Ours (MobileNet) & 1-5   & bs64, ep20 & AffectNet-Top & AffectNetVal-Top & .443 & .568 & .646 & .464 & .496 & .350 & .330 & .321 & .366 & .443 & .443 & .121 \\
    5     & Ours (MobileNet) & 1-5   & bs64, ep20 & AffectNet-Bot. & AffectNetVal-Bot. & .478 & .758 & .362 & .430 & .516 & .450 & .488 & .379 & .442 & .478 & .478 & .124 \\
    \midrule
          &       &       &       &       & \textbf{Test Set FePh} &       &       &       &       &       &       &       &       &       &       &       &  \\
    \midrule
    6     & Ours (MobileNet) & 1-5   & bs64, ep20 & AffectNet-Full & FePh-Full & .271 & .095 & .060 & .743 & .263 & .250 & .166 & -     & .121 & .212 & .243 & .233 \\
    7     & Ours (MobileNet) & 1-5   & bs64, ep20 & AffectNet-Top & FePh-Top & .267 & .011 & .158 & \textbf{.892} & .091 & .152 & .005 & -     & .025 & .167 & .191 & .316 \\
    8     & Ours (MobileNet) & 1-5   & bs64, ep20 & AffectNet-Bot. & FePh-Bot. & .223 & .050 & .006 & .564 & .234 & .286 & .305 & -     & .015 & .182 & .208 & .202 \\
    \midrule
    9     & Desphande & MS, 1, 5 & SAM, bs?, ft3-6 & FePh-Full (5f) & FePh-Full (5f) & .665 & .842 & .678 & .737 & .316 & .682 & .677 & -     & .690 & -     & .661 & .162 \\
    \midrule
    10    & Ours (MobileNet) & 1-5   & SAM, bs32, ft3-6 & FePh-Full (5f) & FePh-Full (5f) & .744 & .899 & .744 & .748 & .662 & .708 & .781 & -     & .774 & .759 & .759 & .073 \\
    11    & Ours (MobileNet) & 1-5   & SAM, bs32, ft3-6 & FePh-Top (5f) & FePh-Top (5f) & .656 & .771 & .653 & .633 & .578 & .669 & .680 & -     & .723 & .672 & .672 & .062 \\
    12    & Ours (MobileNet) & 1-5   & SAM, bs32, ft3-6 & FePh-Bot. (5f) & FePh-Bot. (5f) & .673 & .860 & .696 & .643 & .572 & .652 & .743 & -     & .738 & .700 & .700 & .092 \\
    \midrule
    13    & Ours (MobileNet) & 1-5   & SAM, bs16, ft20-50 & FePh-Full (5f) & FePh-Full (5f) & \textbf{.840} & \textbf{.899} & \textbf{.847} & .870 & \textbf{.772} & \textbf{.810} & \textbf{.855} & -     & \textbf{.813} & \textbf{.838} & \textbf{.838} & \textbf{.042} \\
    14    & Ours (MobileNet) & 1-5   & SAM, bs16, ft20-50 & FePh-Top (5f) & FePh-Top (5f) & .789 & .805 & .805 & .816 & .698 & .802 & .738 & -     & .788 & .779 & .779 & .043 \\
    15    & Ours (MobileNet) & 1-5   & SAM, bs16, ft20-50 & FePh-Bot. (5f) & FePh-Bot. (5f) & .799 & .883 & .813 & .852 & .707 & .739 & .812 & -     & .763 & .796 & .796 & .062 \\
    \bottomrule
    \end{tabular}%
    \caption{Summary of the results for the experiments performing facial expression classification using full face, only top, or only bottom. Image processing codes (sec. \ref{sec:img-pre-proc}): MS=mean subtraction, 1=crop face; 2=square box; 3=zoom out; 4=histogram equalization; 5=rotate. For the hyper-parameters columns: \emph{bs} stands for batch size; \emph{ep} for training epochs; and \emph{ftX-Y} for fine-tuning of further X and then Y epochs, as described in section \ref{sec:train-test}. Here, (5f) indicates 5-fold cross-validation.}
    \label{tab:results}
\end{minipage}
\end{sideways}
\end{table*}

Table \ref{tab:results} reports the results of our experiments, which are commented in the following subsections.
In the table, we report the model architecture, image pre-processing steps, information about the training hyper-parameters, and the training and test sets.
Performances are reported in terms of global accuracy, followed by per-class sensitivity information. The sensitivities are then summarized into balanced accuracy (weighted mean of the per-class sensitivities), and mean of per-class sensitivities (thus disregarding sample imbalance). Finally, we report the standard deviation (SD) among sensitivities, which gives an indication of the reliability of the model across classes.

\subsection{Baselines}

As a baseline and reference method for our work, we use the state of the art method presented by Savchenko \cite{savchenko_facial_2021} (row 1).
It is characterized by an image pre-processing phase that crops the face region and rotates it to position the eyes on the same horizontal level, corresponding to steps 1 and 5 of our pipeline (see section \ref{sec:img-pre-proc}).
This model reaches an overall accuracy of 56.1\% and a balanced accuracy of 56.1\%, but never predicts contempt (under represented class) (SD=0.251).
The image preprocessing contains an RGB centering stage (MS, mean subtraction) that is performed on each input image according to mean RGB values measured in the training set.

\subsection{Experiments on AffectNet}

\begin{figure*}[t]
    \centering
    \includegraphics[width=0.92\textwidth]{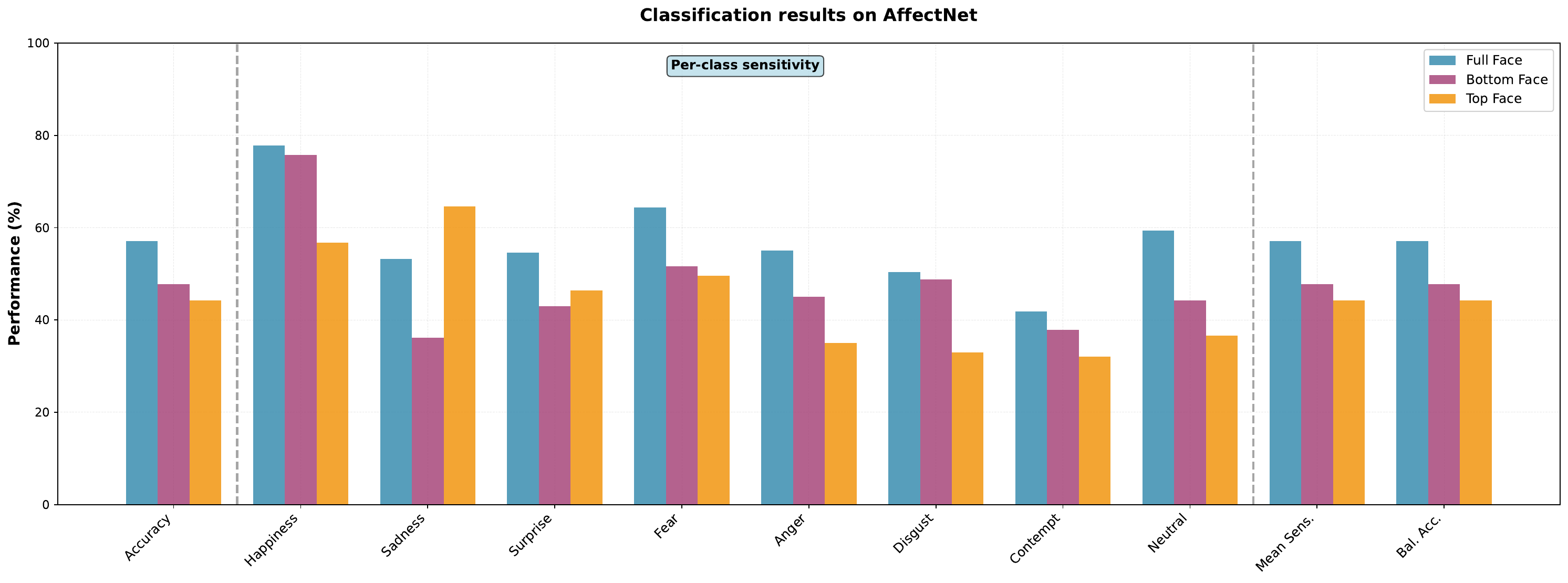}\\
    \includegraphics[width=0.92\textwidth]{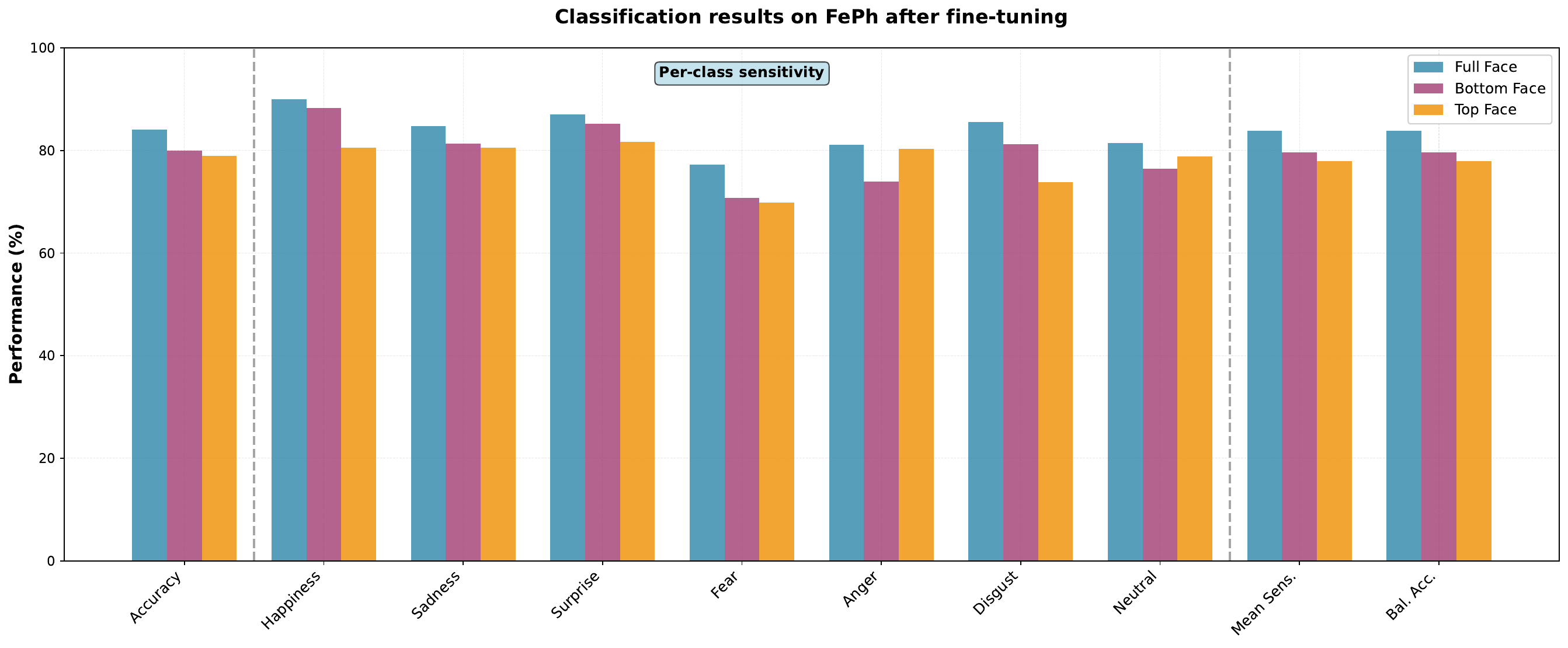}
    \caption{Classification results on the AffectNet dataset (top), and on the FePH dataset (bottom).}
    \label{fig:plot-results}
\end{figure*}

Row 2 of the table shows the results of our reproduction of the Savchenko training pipeline, with the introduction of two image pre-processing stages that square the cropped face (step 2) and zoom-out the cropped region (step 3). We removed the image mean subtraction to set a new reference for our color normalization technique.
This model performs slightly worse than the baseline (mean sensitivity 54.9\%, -1.2\%) but performs more evenly among classes (SD=0.113): contempt is now better classified, but there is a performance loss for happiness, surprise, fear, and neutral.\\

In the next tests, we introduced the Histogram Equalization. Details are reported in rows 3--5 of table \ref{tab:results} and plotted in figure~\ref{fig:plot-results}~(top).

Row 3 shows that this training pipeline has a slightly higher mean sensitivity (57.1\%) than the state of the art (+0.9\%) and our training without histogram equalization (+2.2\%). In addition, the SD among per-class sensitivity has improved (SD=0.106).
With respect to the baseline, this method also removes hard-coded RGB means from the implementation and is supposed to generalize better to other datasets like FePh.

Finally, the results of the tests with partially masked faces (rows 4--5) give a reference of how well the machine learning model can infer facial expression using only partially visible information.
By using only the upper part of the face, mean sensitivity is as high as 44.3\%, while by using the lower part of the face it reaches 47.8\%.
This suggests that by using only half of the face, it is already possible to infer quite reliably the expression of an individual, far above the pure chance level of 1/8 (12.5\%)


\subsection{Experiments on FePh}

For the tests on the FePh dataset, the results are reported in rows 6--15 of table \ref{tab:results}.
When simply testing the FePh dataset on the model trained on the AffectNet dataset (rows 6--8), results show a mean sensitivity as low as 24.3\%, with a clear bias toward surprise.

To improve performance, we applied the fine-tuning procedure described by Deshpande et al.~\cite{deshpande_fine-tuning_2022} (row 9), which applies a fine-tuning on the FePH dataset in two stages: 3 epochs training only the classification layers, and another 6 epochs fine-tuning the whole model. However, the image pre-processing stage uses the mean subtraction with values already provided by Savchenko. The batch size is not known.
The original paper~\cite{deshpande_fine-tuning_2022} reports a mean sensitivity of 66.1\%, but a low performance on the Fear expression is noticeable.\\

We thus reproduced the training procedure introducing the Histogram Equalization in the pre-processing pipeline and setting the batch size to 32 (rows 10--12). This led to an increase of performance to 75.9\% mean sensitivity, and an SD as low as 0.073.
For this configuration, the ability to recognize emotions using only the upper/lower part of the face increases to 67.2/70.0\%.\\

With a further grid search exploration of the training hyper-parameters, we found the best combination with a batch size of 16 and longer fine-tuning stages of 20 and 50 epochs (rows 13--15).
The results of our best model on FePh are plotted in figure \ref{fig:plot-results} (bottom). They show a mean sensitivity of 83.8\%, with a SD of only 0.042.
In this case, the mean sensitivities in recognizing facial expressions only with the upper and lower parts of the face increase to 77.9\% and 79.6\%, respectively.

\section{Results and discussion}

The results on the FePh dataset show that with proper color normalization and longer fine-tuning it is possible to recognize facial expressions on the FePh dataset with an average sensitivity of 83.8\%, representing a +26\% w.r.t. our baseline methods.
Remarkably, the mean sensitivity when recognizing facial expressions from only the upper or lower parts of the face (77.9\%, 79.6\%) is also: i) considerably higher than using the full face on the baseline methods, and ii) only marginally lower (-5.9\%, -4.2\%) than using the full face.

In general, in line with the work presented in the introduction, the recognition from the bottom part of the face is higher than that from the top part, confirming previous findings in the literature on human performances.

A comparison with human performances reported by Letourneau et al.~\cite{letourneau_gaze_2011} is summarized in table \ref{tab:comparison-with-humans}. Humans are better at recognizing facial expressions from the entire face and the bottom part (+10.54\%, +8.39\%). However, the neural network can more reliably recognize emotions from the top part of the face (+9.56\%).

\begin{table}
    \centering
    \scriptsize
\begin{tabular}{ll|lll}
\toprule
      &       & \multicolumn{3}{c}{\textbf{Face}} \\
\textbf{Source} & \textbf{Classifier} & \textbf{Full} & \textbf{Top} & \textbf{Bot.} \\
\midrule
Our model & MobileNetV2 & 83.8  & \textbf{77.9} & 79.6 \\
Letourneau et al. & Hearing humans & \textbf{94.34} & 68.34 & \textbf{91.89} \\
Letourneau et al. & Deaf humans & 93.07 & 60.17 & 85.93 \\
\bottomrule
\end{tabular}%
    \caption{Comparison of classification accuracy between our machine learning model and the human performance reported by Letourneau et al.~\cite{letourneau_gaze_2011}.}
    \label{tab:comparison-with-humans}
\end{table}


In general, on the FePh dataset, all emotions are recognized better from the bottom part than from the top, with the exception of anger, probably due to the extreme activation of the forehead muscle characterizing this emotion.
However, this is not consistent with the behavior in the AffectNet dataset, where the expressions better recognized from top than bottom are sadness and surprise, possibly because of the characteristic shape taken by the eyes.

So, we might argue that when fine-tuning for FePh, the attention of the neural network shifts from eye shapes to wrinkle formation, but this is only a seminal hypothesis and needs further investigation using explainable AI methods (such as GradCAM saliency maps \cite{selvaraju_grad-cam_2017}) to be confirmed.

\section{Limitations}

This work also presents some limitations with respect to research in the sign language domain.

First of all, the interpreters of the PHOENIX dataset are not native signers, and they are not performing natural interaction, but staged formalized communications in front of a camera. These two aspects prevent us from generalizing our observation to native speakers in casual communication.

Second, the PHOENIX dataset has a very specific color profile and pretty low resolution images (210x300 pixels), with face crops often resulting in very low resolution as about 50x50 pixels.
Hence, while we showed that through a careful pre-processing and fine-tuning procedure it is possible to perform reliable FER on the PHOENIX dataset, the fine-tuning process most probably adapts to the image quality and resolution of the FePh dataset rather than really grasping a different way of expressing emotions from signers.

\section{Conclusions}

We presented a method to reliably recognize facial expressions from full and half faces on a sign language dataset.

By combining an image pre-processing stage (cropping, squaring, zooming, equalizing color histogram, and rotating) with a double-step fine-tuning procedure, it was possible to reach considerable performances on the FePh dataset: a dataset characterized by a very low resolution and a peculiar color profile.
This result set a baseline for future machine-learning-supported investigations on the use of (upper/lower) facial expressions in sign language communication.

More in general, these results suggest that emotion recognition can also be reliably applied in scenarios of partially occluded faces, such as when subjects are wearing AR/VR glasses.

Further, as the FePh is a subset of PHOENIX, the same image pre-processing and fine-tuning pipeline could help improve any computer vision task involving the PHOENIX dataset, such as sign language recognition.

Future work might investigate whether this pre-processing + double-fine-tuning approach could help improving classification in any other dataset characterized by very specific recording conditions, and thus with low variance in the images color profile, which is often the case for datasets recorded in television broadcasters' production studios.

%
%
%
\section*{Acknowledgments}
This research was funded by the German Ministry of Education and Research (BMBF) through the AVASAG project (grant number 16SV8491) \cite{nunnari2021AT4SSL-AVASAG,bernhard22PETRA-AVASAG} and through the BIGEKO project (grant number 16SV9093).

\bibliographystyle{plain}
\bibliography{SLTAT2025-TopBottomExpression}

\end{document}